\documentclass{vgtc}                          

\graphicspath{{figures/}{pictures/}{images/}{./}} 

\usepackage{times}                     

\usepackage{tabu}                      
\usepackage{booktabs}                  
\usepackage{lipsum}                    
\usepackage{mwe}                       
\usepackage{algorithm}                 
\usepackage{algorithmicx}
\usepackage{algcompatible}
\usepackage{algpseudocode}
\usepackage{amsmath}                
\usepackage{placeins}
\usepackage{balance}
\usepackage{mathptmx}                  

\onlineid{0}

\vgtccategory{Research}

\vgtcinsertpkg

\title{GeminiPainter: Real-Time AI-Guided Robotic Sketch Generation for Live Portraiture}

\author{Miguel Altamirano Cabrera\thanks{E-mail: m.altamirano@skoltech.ru} %
\and Aleksey Fedoseev\thanks{E-mail: Aleksey.Fedoseev@skoltech.ru} %
\and Iana Zhura\thanks{E-mail: Iana.Zhura@skoltech.ru} %
\and Dzmitry Tsetserukou\thanks{E-mail: d.tsetserukou@skoltech.ru}%
}
\affiliation{\scriptsize Intelligent Space Robotics Lab, Skolkovo Institute of Science and Technology, Moscow, Russian Federation}

\teaser{
  \centering
  \makebox[\textwidth][c]{%
    \hspace*{0.2cm}%
    \includegraphics[width=1.18\textwidth]{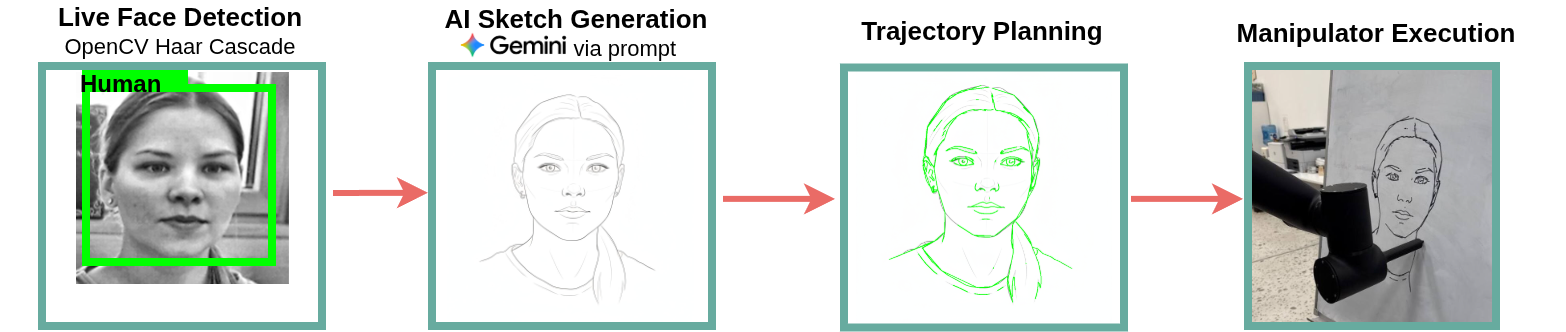}
    
  }
  \caption{GeminiPainter pipeline comprising live face detection, AI-based sketch generation, trajectory planning, and manipulator execution.}
  \label{fig:pipeline}}

\abstract{

We present an autonomous robotic portrait-generation system combining real-time face detection, AI-based sketch generation, and robotic drawing. The system captures video frames, extracts facial regions, converts them into minimalist single-line sketches using the Gemini Vision API, optimizes stroke order through graph-based path planning, and executes smooth trajectories on a 6-DoF collaborative manipulator. This perception-cognition-action pipeline integrates computer vision, neural artistic abstraction, motion optimization, and robot control. User ratings on a 5-point scale were high for sketch quality ($4.33$), perceived execution ($4.53$), and user experience ($4.65$), indicating recognizable, appealing, and engaging robotic portraits.

} 


\begin{document}


\firstsection{Introduction}

\maketitle


The intersection of artificial intelligence, computer vision, and robotics has opened new possibilities for creative human-robot interaction. While robots have long been employed for manufacturing and precise manipulation, recent advances in deep learning have enabled robots to perform increasingly nuanced creative tasks. Portrait drawing is a compelling application because it combines facial-feature interpretation, artistic abstraction, and precise motion control.

Prior work in robotic drawing has primarily focused on two paradigms: (1) reproducing predesigned artwork through inverse kinematics and trajectory planning, and (2) generating drawings from predefined stylization algorithms \cite{Deussen2000FloatingPA}. Systems such as the drawing robot by Liu et al.~\cite{liuDrawing21} employed image processing pipelines (edge detection and edge thinning algorithms) to extract strokes from photographs. However, purely edge-based approaches can produce geometric and noise-sensitive abstractions with limited semantic stylization.


The emergence of generative models and vision transformers has shifted this paradigm. Text-to-image and multimodal systems, such as DALL-E \cite{ramesh2021zero}, Stable Diffusion \cite{rombach2022high}, and more recently Google's Gemini \cite{geminiteam2025geminifamilyhighlycapable}, can synthesize diverse, semantically faithful imagery from text and image inputs. Our work leverages Gemini’s native image-generation capability to produce minimalist single-line sketches that preserve salient facial features while remaining simple enough for robotic execution.


Prior HRI research has also explored expressive and creative robot interaction through distributed teleoperation \cite{zoomtouch}, wearable motion-based cobot control \cite{cobotgear}, aerial spray painting \cite{DroneGraffiti}, projected interaction \cite{LightAir}, and long-exposure light painting \cite{DroneLight}. GeminiPainter extends this broader line of work by combining generative visual abstraction with autonomous trajectory execution on a collaborative manipulator.

Real-time face detection has matured significantly with advances in efficient neural networks. MediaPipe Face Detection \cite{lugaresi2019mediapipe}, YOLO-Face \cite{yu2022yolofacev2scaleocclusionaware}, and RetinaFace \cite{deng2019retinafacesinglestagedenseface} offer fast, accurate detection with low latency suitable for interactive applications. These systems enable practical facial region-of-interest extraction, essential for our pipeline's first perception stage.

The problem of ordering drawing strokes to minimize ``lift distance'' (non-drawing movements) is analogous to the traveling salesman problem. Classical approaches include greedy nearest-neighbor methods and local-search heuristics such as Lin–Kernighan optimization \cite{helsgaun2000effective}. Our system applies stroke ordering to reduce non-drawing travel and potentially shorten portrait-execution time.

Collaborative robots (cobots) like the 6-DoF ARM-95 collaborative robot from Applied Robotics, support close-proximity human-robot operation through controller-level safety limits, compliant behavior, and smooth trajectory execution. Modern controllers support trajectory blending, wherein intermediate waypoints are connected via smooth circular arcs rather than point-to-point linear motions, enabling continuous, high-speed operation \cite{hoganimpedance}. This capability is central to our smooth execution mode.

While individual components (face detection, sketch generation, robot control) are well-established, their integration into a cohesive, real-time interactive system for live portrait generation presents several novel aspects. The main contribution of this work is an integrated robotic portrait-drawing pipeline that combines: (1) VLM-based semantic sketch abstraction, (2) heuristic stroke ordering with local path refinement, and (3) blended Cartesian trajectory execution on a collaborative manipulator. We additionally provide an exploratory characterization of computational latency and user perceptions in a study with 10 participants.


\section{Methods}

\subsection{System Architecture}

The GeminiPainter pipeline is shown in (Fig.~\ref{fig:pipeline}).

\begin{figure*}[h]
 \centering 

 \includegraphics[width=1.9\columnwidth]{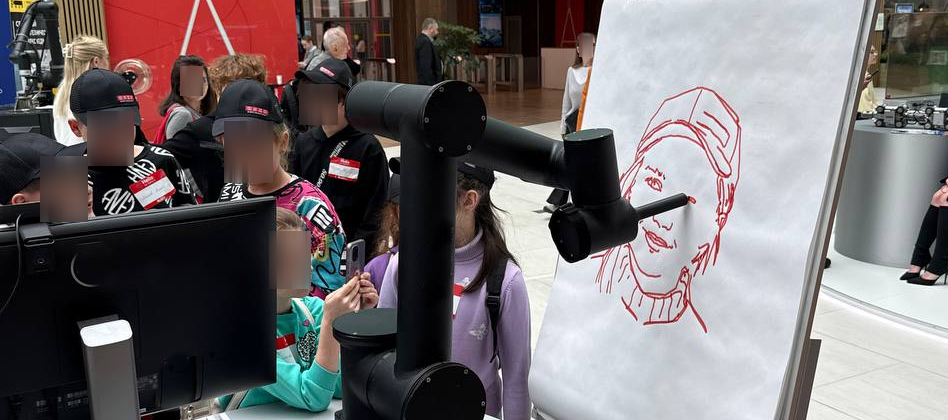}
 \caption{GeminiPainter generating and drawing a portrait during a live demonstration using VLM-based sketch generation and optimized stroke ordering.}
 \label{fig:teaser}
\end{figure*}

The pipeline is modular: individual perception, generation, and planning components can be replaced independently, and the execution module can be adapted to different robot models. The system configuration utilized in this work consists of the following subsystems:
\begin{enumerate}
    \item \textbf{Live Face Detection:} acquisition of a camera frame and extraction of the facial region of interest.
    \item \textbf{Sketch Generation:} conversion of the facial crop into a minimalist line sketch using the Gemini API.
    \item \textbf{Trajectory Planning:} extraction, simplification, ordering, and coordinate mapping of drawing strokes.
    \item \textbf{Manipulator Execution:} inverse-kinematics computation and blended trajectory execution.
\end{enumerate}

\subsection{Live Face Detection}

We perform real-time face detection on standard USB camera input ($30$ FPS) using a Haar feature-based cascade classifier \cite{990517}, which provides low-latency frontal-face detection without requiring GPU acceleration. We select the largest detected face bounding box, crop it with a small margin, and apply Contrast-Limited Adaptive Histogram Equalization (CLAHE) to normalize contrast under varying lighting. The enhanced face crop $\mathbf{F}$ is passed to sketch generation.

\subsection{AI-Driven Sketch Generation}

Traditional edge detection (Canny, Sobel) produces geometric, brittle abstractions sensitive to noise and lighting. Large vision models can learn semantically meaningful visual features and stylistic conventions from diverse image-text data. Recent advances in diffusion models \cite{ho2020denoising} have demonstrated remarkable capabilities in conditional image generation, which we leverage for sketch synthesis.

We prompt the Gemini-2.5-flash-imag model through the Gemini Vision API with a text prompt specifically engineered to emphasize continuous lines:

\textit{``Generate a minimalist, single-line portrait sketch of this face. Draw only essential features (eyes, nose, mouth, jawline). Use a single continuous or near-continuous line. Output should resemble a line drawing suitable for robotic drawing. Minimize the number of strokes.''}

The prompt constrains the output along two dimensions: semantic prioritization and physical simplicity. First, limiting the portrait to essential facial features encourages the model to omit shading and texture, reducing downstream image-processing complexity. Second, requesting a continuous or near-continuous line with few strokes reduces the number of pen-up transitions and simplifies stroke ordering and robotic execution. The model returns a PNG image of the generated sketch. 

\begin{algorithm}
\caption{Sketch to Stroke Extraction}
\begin{algorithmic}
\State $\mathbf{S} \gets \text{PNG2Gray}(\text{GeneratedSketch})$
\State $\mathbf{B} \gets \text{Threshold}(\mathbf{S}, \tau_{\text{luminance}})$
\State $\text{Contours} \gets \text{FindContours}(\mathbf{B})$
\For{each contour $c$ in Contours}
    \If{Area$(c) > A_{\min}$}
        \State $\mathbf{S}_i \gets \text{RamerDouglasPeucker}(c, \epsilon)$
        \State Append $\mathbf{S}_i$ to StrokeList
    \EndIf
\EndFor
\State \Return StrokeList
\end{algorithmic}
\end{algorithm}

We then extract stroke coordinates using the image-processing procedure summarized in Algorithm 1. The resulting stroke set is denoted by $\mathbf{S}_i=\{\mathbf{S}_1, \mathbf{S}_2, \ldots, \mathbf{S}_n\}$, where each stroke $\mathbf{S}_i = [\mathbf{p}_1, \mathbf{p}_2, \ldots, \mathbf{p}_{m_i}]$ is an ordered sequence of 2D points.

\subsection{Trajectory Planning}
\subsubsection{Stroke Ordering and Path Optimization}

Drawing efficiency is influenced by the non-drawing travel between consecutive stroke endpoints. We formulate stroke ordering as a variant of the traveling-salesman path problem.:

\begin{equation}
\min_{\pi} \sum_{i=1}^{n-1} d(\mathbf{p}_{\text{end}}(\mathbf{S}_{\pi(i)}), \mathbf{p}_{\text{start}}(\mathbf{S}_{\pi(i+1)})),
\end{equation}
where $\pi$ is the permutation of $\{1, 2, \ldots, n\}$, and $d(\cdot, \cdot)$ is the Euclidean distance.

We employ a greedy nearest-neighbor heuristic with post-optimization:

\begin{algorithm}
\caption{Stroke Ordering}
\begin{algorithmic}
\State $\text{Order} \gets []$, $\text{Remaining} \gets \{1, 2, \ldots, n\}$
\State $\text{Current} \gets \mathbf{S}_1$, Remaining $\gets$ Remaining $\setminus \{1\}$
\While{Remaining $\neq \emptyset$}
    \State $j^* \gets \arg\min_j d(\mathbf{p}_{\text{end}}(\text{Current}), \mathbf{p}_{\text{start}}(\mathbf{S}_j))$ for $j \in$ Remaining
    \State Order $\gets$ Order $\cup \{j^*\}$
    \State Current $\gets \mathbf{S}_{j^*}$, Remaining $\gets$ Remaining $\setminus \{j^*\}$
\EndWhile
\State \Return Order
\end{algorithmic}
\end{algorithm}

Additionally, we check if reversing any stroke reduces distance and apply local 2-opt swaps.

\subsubsection{Coordinate Mapping}

Face images are captured in pixel coordinates $[u, v]$ with origin at image top-left. Whiteboard positions are defined in 3D robot coordinates (meters) relative to the robot base frame. We perform affine mapping which is calibrated via four-point correspondence (image corners to whiteboard corners). We maintain separate lift and drawing Z-offsets to ensure the tool (marker/pen) makes contact only during drawing.

\subsection{Manipulator Execution}
\subsubsection{End-Effector Design and Marker Holder Mechanism}


The end-effector integrates a custom marker holder with a passive mechanical spring mechanism ($1.5\,\text{cm}$ of vertical compliance). This compliance accommodates surface non-planarity, dampens contact impact during tool-transition lifts, and helps maintain consistent marker contact across the drawing surfac across the whiteboard canvas.

\subsubsection{Trajectory Execution}

Trajectory execution consists of two stages: inverse-kinematics computation for the Cartesian drawing poses and time-parameterized trajectory interpolation with waypoint blending.





This approach maintains continuous, high-speed motion while preserving geometric accuracy of stroke features.

The execution protocol proceeds stroke by stroke. Starting from a retracted pose at the clearance Z-offset, the tool moves to the stroke's start point via a blended trajectory, lowers to the drawing surface, and traces the stroke waypoints with trajectory blending. It then retracts to the clearance height and advances to the start of the next stroke, repeating until all strokes are drawn.

Safety constraints include joint velocity/acceleration limits, singularity avoidance, and emergency abort triggers.

\section{Experimental Validation}

We report a user study ($N=10$ participants) that evaluates the system along two dimensions: (i) the computational latency of the perception-generation-planning pipeline and (ii) participants’ perceptions of the generated sketches, robotic rendering, and overall interaction.

\subsection{Experimental Design}

\textbf{Participants and apparatus.} Ten participants (3 women and 7 men, aged 23--37) completed one session each. All participants provided informed consent for participation and facial-image processing. For each session, the system captured a live frame, detected and cropped the face, generated a single-line portrait via the Gemini Vision API, extracted and ordered the strokes, mapped them to robot coordinates, and rendered the resulting drawing. Stage-by-stage timing was logged automatically. 

\textbf{Measures.} We recorded two types of measures. First, we logged the processing latency $L = t_{\text{face}} + t_{\text{gemini}} + t_{\text{stroke}} + t_{\text{planning}} + t_{\text{IK}}$, representing the end-to-end computational budget required to prepare motion trajectories prior to physical execution, relative to a design target of $L < 10\,\text{s}$. Second, after each session participants rated ten items on a 5-point Likert scale (1 = strongly disagree, 5 = strongly agree), grouped into three constructs: \emph{sketch quality} (recognizability, artistic quality, essential features), \emph{perceived execution} (line continuity, line-width consistency, feature closure), and \emph{user experience} (overall impression, feedback clarity, perceived autonomy, engagement intent).

\subsection{Results}

\subsubsection{Real-Time Performance}

Table~\ref{tab:latency} reports the latency breakdown. Mean pipeline latency before physical robot motion was $8.5 \pm 2.1$\,s (range $6.6$--$13.8$\,s). The mean was below the 10 s design target, although some trials exceeded it. The remote Gemini image-generation call dominated the measured latency, accounting for $98.5\%$ of total latency ($8.38 \pm 2.07$\,s), while \emph{all} on-device computation, face detection ($30$\,ms), stroke extraction ($82$\,ms), path planning ($9$\,ms), and inverse kinematics ($6$\,ms), together required only $\approx 127$\,ms ($1.5\%$). Local processing therefore imposes negligible overhead, and end-to-end responsiveness is determined almost entirely by the latency of the remote vision model. While physical robot actuation adds execution time depending on path length, maintaining a sub-10\,s computation budget ($8.5 \pm 2.1\,\text{s}$) is critical for live HRI, ensuring the user is not left idling during perception and trajectory synthesis.

\begin{table}[t]
\centering
\footnotesize
\caption{Pipeline latency breakdown over $N=10$ runs. The Gemini image-generation call dominates end-to-end latency.}
\label{tab:latency}
\begin{tabular}{lrrr}
\toprule
Stage & Mean (ms) & SD (ms) & \% of total \\
\midrule
Face detection & 30.4 & 28.0 & 0.4 \\
Gemini API & 8377.7 & 2066.3 & 98.5 \\
Stroke extraction & 82.4 & 49.7 & 1.0 \\
Path planning & 8.6 & 7.3 & 0.1 \\
IK computation & 5.7 & 1.2 & 0.1 \\
\midrule
\textbf{Total} & 8504.8 & 2061.7 & 100.0 \\
\bottomrule
\vspace{-3mm}
\end{tabular}
\end{table}

\subsubsection{User Evaluation}

All ten survey items had mean ratings above the neutral midpoint of 3 (Table~\ref{tab:likert}). At the construct level, \emph{user experience} was rated highest (mean $4.65 \pm 0.46$), followed by \emph{perceived execution} ($4.53 \pm 0.63$) and \emph{sketch quality} ($4.33 \pm 0.70$). The strongest individual items were engagement intent ($4.80 \pm 0.42$) and feedback clarity ($4.70 \pm 0.48$), suggesting that participants found the system compelling and the live camera/detection feedback easy to follow. Recognizability showed the widest dispersion ($4.30 \pm 1.06$): most participants rated their portraits as recognizable, although ratings varied substantially across participants.

\begin{table}[t]
\centering
\footnotesize
\caption{User ratings ($N=10$) on a 5-point Likert scale (1 = strongly disagree, 5 = strongly agree).}
\label{tab:likert}
\begin{tabular}{llrr}
\toprule
Construct & Item & Mean & SD \\
\midrule
Sketch Quality & Recognizability & 4.30 & 1.06 \\
 & Artistic quality & 4.30 & 0.82 \\
 & Essential features & 4.40 & 0.84 \\
\midrule
Perceived Execution & Line continuity & 4.40 & 0.97 \\
 & Line-width consistency. & 4.50 & 0.71 \\
 & Feature closure & 4.70 & 0.48 \\
\midrule
User Experience & Overall impression & 4.60 & 0.97 \\
 & Feedback clarity & 4.70 & 0.48 \\
 & Perceived autonomy & 4.50 & 0.71 \\
 & Engagement intent & 4.80 & 0.42 \\
\midrule
\multicolumn{2}{l}{\textit{Sketch quality (mean)}} & 4.33 & 0.70 \\
\multicolumn{2}{l}{\textit{Perceived execution (mean)}} & 4.53 & 0.63 \\
\multicolumn{2}{l}{\textit{User experience (mean)}} & 4.65 & 0.46 \\
\bottomrule
\vspace{-9mm}
\end{tabular}
\end{table}

\subsection{Discussion}

A potential advantage of integrating a VLM instead of relying exclusively on conventional edge detection is semantic abstraction. Conventional edge detectors operate on local intensity gradients and can produce fragmented or noise-sensitive contours under challenging imaging conditions. In contrast, prompting Gemini to generate a sparse line portrait delegates high-level feature selection to the generative model. This can simplify subsequent vectorization and path planning. However, because the present evaluation does not include an edge-detection or local-generation baseline, the relative benefits of the proposed approach remain to be established experimentally.


Rather than substituting human artistic mastery, GeminiPainter serves as an interactive performative medium where real-time computational abstraction and robotic actuation converge to engage users in live co-presence.

\section{Conclusion}

We investigate an autonomous robotic portrait generation system that combines real-time computer vision, deep neural networks for sketch abstraction, path planning algorithms, and collaborative robotic control to enable autonomous portrait drawing. Key technical contributions include: (1) the application of large vision models (Gemini Vision) for semantically meaningful sketch generation, (2) efficient combinatorial optimization of stroke sequences to minimize non-productive motion, and (3) trajectory blending strategies that enable smooth, continuous robot operation. Collectively, these techniques demonstrate that modern deep learning and collaborative manipulators can execute nuanced artistic tasks that traditionally required human skill.

The system architecture integrates digital perception (face detection), AI-driven abstraction (Gemini sketch generation), and physical robotic execution (6-DoF collaborative ARM-95 manipulator) within a structured human-robot interaction context. The mechanical end-effector design—featuring a compliant marker holder with $1.5$ cm vertical compliance—enables robust sketch rendering despite surface irregularities. Real-time feedback mechanisms and interactive user guidance establish a dynamic relationship between human subjects and robotic agents during collaborative artwork generation.


Future work includes: (1) conducting comparative baseline evaluations against classical edge-based vectorization and unoptimized stroke orderings, (2) extending to multi-face collaborative drawing scenarios, (3) incorporating adaptive drawing speed and orientation strategies to achieve varied line weights, (4) exploring alternative sketch models (fine-tuned diffusion models and local VLMs), and (5) deploying the system in public contexts to study long-term user engagement.

\section*{Acknowledgements} 
Research reported in this publication was financially supported by the RSF grant No. 24-41-02039.
\balance


\begin{thebibliography}{10}

\bibitem{deng2019retinafacesinglestagedenseface}
J.~Deng, J.~Guo, Y.~Zhou, J.~Yu, I.~Kotsia, and S.~Zafeiriou.
\newblock {RetinaFace}: Single-stage dense face localisation in the wild, 2019.
\newblock arXiv:1905.00641.

\bibitem{Deussen2000FloatingPA}
O.~Deussen, S.~Hiller, C.~Van~Overveld, and T.~Strothotte.
\newblock Floating points: A method for computing stipple drawings.
\newblock {\em Computer Graphics Forum}, 19:41--50, 2000. doi: {{%
10\hspace{.1pt}\discretionary{.}{%
}{.}\hspace{.4pt}1111\discretionary{/}{%
}{/}1467\discretionary{%
}{-}{-}8659\hspace{.1pt}\discretionary{.}{%
}{.}\hspace{.4pt}00396}}


\bibitem{helsgaun2000effective}
K.~Helsgaun.
\newblock An effective implementation of the lin–kernighan traveling salesman heuristic.
\newblock {\em European Journal of Operational Research}, 126(1):106--130, 2000. doi: {{%
10\hspace{.1pt}\discretionary{.}{%
}{.}\hspace{.4pt}1016\discretionary{/}{%
}{/}S0377\discretionary{%
}{-}{-}2217\discretionary{%
}{(}{(}99\discretionary{)}{%
}{)}00284\discretionary{%
}{-}{-}2}}


\bibitem{cobotgear}
J.~Heredia, M.~Altamirano~Cabrera, J.~Tirado, V.~Panov, and D.~Tsetserukou.
\newblock {CobotGear}: Interaction with collaborative robots using wearable optical motion capturing systems.
\newblock In {\em Proc. IEEE Int. Conf. on Automation Science and Engineering (CASE)}, pp. 1584--1589, 2020. doi: {{%
10\hspace{.1pt}\discretionary{.}{%
}{.}\hspace{.4pt}1109\discretionary{/}{%
}{/}CASE48305\hspace{.1pt}\discretionary{.}{%
}{.}\hspace{.4pt}2020\hspace{.1pt}\discretionary{.}{%
}{.}\hspace{.4pt}9217041}}


\bibitem{ho2020denoising}
J.~Ho, A.~Jain, and P.~Abbeel.
\newblock Denoising diffusion probabilistic models.
\newblock In {\em Proc. of the Int. Conf. on Neural Information Processing Systems (NIPS)}, pp. 6840--6851, 2020.

\bibitem{hoganimpedance}
N.~Hogan.
\newblock Impedance control: An approach to manipulation.
\newblock In {\em Proc. American Control Conf.}, pp. 304--313, 1984. doi: {{%
10\hspace{.1pt}\discretionary{.}{%
}{.}\hspace{.4pt}23919\discretionary{/}{%
}{/}ACC\hspace{.1pt}\discretionary{.}{%
}{.}\hspace{.4pt}1984\hspace{.1pt}\discretionary{.}{%
}{.}\hspace{.4pt}4788393}}


\bibitem{DroneLight}
R.~Ibrahimov, N.~Zherdev, and D.~Tsetserukou.
\newblock Dronelight: Drone draws in the air using long exposure light painting and ml.
\newblock In {\em Proc. IEEE Int. Conf. on Robot and Human Interactive Communication (RO-MAN)}, pp. 446--450, 2020. doi: {{%
10\hspace{.1pt}\discretionary{.}{%
}{.}\hspace{.4pt}1109\discretionary{/}{%
}{/}RO\discretionary{%
}{-}{-}MAN47096\hspace{.1pt}\discretionary{.}{%
}{.}\hspace{.4pt}2020\hspace{.1pt}\discretionary{.}{%
}{.}\hspace{.4pt}9223601}}


\bibitem{liuDrawing21}
F.~Liu, L.~Cao, Z.~Sun, and Z.~Li.
\newblock Research on drawing robot based on image edge detection.
\newblock In {\em Proc. Int. Conf. on Control, Robotics and Intelligent System (CCRIS)}, p. 6–11, 2021. doi: {{%
10\hspace{.1pt}\discretionary{.}{%
}{.}\hspace{.4pt}1145\discretionary{/}{%
}{/}3483845\hspace{.1pt}\discretionary{.}{%
}{.}\hspace{.4pt}3483847}}


\bibitem{lugaresi2019mediapipe}
C.~Lugaresi et~al.
\newblock Mediapipe: A framework for building perception pipelines, 2019.
\newblock arXiv:1906.08172.

\bibitem{LightAir}
M.~Matrosov, O.~Volkova, and D.~Tsetserukou.
\newblock Lightair: a novel system for tangible communication with quadcopters using foot gestures and projected image.
\newblock In {\em Proc. ACM SIGGRAPH Emerging Technologies}, pp. 1--2, 2016. doi: {{%
10\hspace{.1pt}\discretionary{.}{%
}{.}\hspace{.4pt}1145\discretionary{/}{%
}{/}2929464\hspace{.1pt}\discretionary{.}{%
}{.}\hspace{.4pt}2932429}}


\bibitem{ramesh2021zero}
A.~Ramesh et~al.
\newblock Zero-shot text-to-image generation.
\newblock In {\em Proc. of the Int. Conf. on Machine Learning (ICML)}, pp. 8821--8831. PMLR, 2021.

\bibitem{rombach2022high}
R.~Rombach, A.~Blattmann, D.~Lorenz, P.~Esser, and B.~Ommer.
\newblock { High-Resolution Image Synthesis with Latent Diffusion Models}.
\newblock In {\em Proc. IEEE/CVF Conf. on Computer Vision and Pattern Recognition (CVPR)}, pp. 10674--10685, 2022. doi: {{%
10\hspace{.1pt}\discretionary{.}{%
}{.}\hspace{.4pt}1109\discretionary{/}{%
}{/}CVPR52688\hspace{.1pt}\discretionary{.}{%
}{.}\hspace{.4pt}2022\hspace{.1pt}\discretionary{.}{%
}{.}\hspace{.4pt}01042}}


\bibitem{geminiteam2025geminifamilyhighlycapable}
G.~Team et~al.
\newblock Gemini: A family of highly capable multimodal models, 2025.
\newblock arXiv:2312.11805.

\bibitem{DroneGraffiti}
A.~Uryasheva, M.~Kulbeda, N.~Rodichenko, and D.~Tsetserukou.
\newblock Dronegraffiti: autonomous multi-uav spray painting.
\newblock In {\em Proc. ACM SIGGRAPH Studio}, pp. 1--2, 2019. doi: {{%
10\hspace{.1pt}\discretionary{.}{%
}{.}\hspace{.4pt}1145\discretionary{/}{%
}{/}3306306\hspace{.1pt}\discretionary{.}{%
}{.}\hspace{.4pt}3328000}}


\bibitem{990517}
P.~Viola and M.~Jones.
\newblock Rapid object detection using a boosted cascade of simple features.
\newblock In {\em Proc. IEEE Conf. on Computer Vision and Pattern Recognition (CVPR)}, vol.~1, pp. 1--9, 2001. doi: {{%
10\hspace{.1pt}\discretionary{.}{%
}{.}\hspace{.4pt}1109\discretionary{/}{%
}{/}CVPR\hspace{.1pt}\discretionary{.}{%
}{.}\hspace{.4pt}2001\hspace{.1pt}\discretionary{.}{%
}{.}\hspace{.4pt}990517}}


\bibitem{yu2022yolofacev2scaleocclusionaware}
Z.~Yu, H.~Huang, W.~Chen, Y.~Su, Y.~Liu, and X.~Wang.
\newblock {YOLO-FaceV2}: A scale and occlusion aware face detector.
\newblock {\em Pattern Recognition}, pp. 1--10, Nov. 2024. doi: {{%
10\hspace{.1pt}\discretionary{.}{%
}{.}\hspace{.4pt}1016\discretionary{/}{%
}{/}j\hspace{.1pt}\discretionary{.}{%
}{.}\hspace{.4pt}patcog\hspace{.1pt}\discretionary{.}{%
}{.}\hspace{.4pt}2024\hspace{.1pt}\discretionary{.}{%
}{.}\hspace{.4pt}110714}}


\bibitem{zoomtouch}
I.~Zakharkin, A.~Tsaturyan, M.~Altamirano~Cabrera, J.~Tirado, and D.~Tsetserukou.
\newblock Zoomtouch: Multi-user remote robot control in zoom by dnn-based gesture recognition.
\newblock In {\em Proc. SIGGRAPH Asia Emerging Technologies}, pp. 1--2, 2020. doi: {{%
10\hspace{.1pt}\discretionary{.}{%
}{.}\hspace{.4pt}1145\discretionary{/}{%
}{/}3415255\hspace{.1pt}\discretionary{.}{%
}{.}\hspace{.4pt}3422892}}


\end{thebibliography}
\end{document}